%% file: PaperForReview.tex
\documentclass[10pt,twocolumn,letterpaper]{article}

\usepackage[pagenumbers]{cvpr} 

\usepackage[accsupp]{axessibility}  
\usepackage{graphicx}
\usepackage{amsmath}
\usepackage{amssymb}
\usepackage{booktabs}
\usepackage{multirow}
\usepackage{tabularx}
\usepackage{pifont}
\usepackage[misc]{ifsym}

\newcommand{\tf}[1]{{\color{red} [ToFill]}}
\newcommand{\tc}[1]{{\color{red} [ToCite]}}

\makeatletter
\def\blfootnote{\xdef\@thefnmark{}\@footnotetext}
\makeatother

\usepackage[pagebackref,breaklinks,colorlinks]{hyperref}

\usepackage[capitalize]{cleveref}
\crefname{section}{Sec.}{Secs.}
\Crefname{section}{Section}{Sections}
\Crefname{table}{Table}{Tables}
\crefname{table}{Tab.}{Tabs.}

\def\cvprPaperID{7436} 
\def\confName{CVPR}
\def\confYear{2023}

\begin{document}

\title{Fine-grained Audible Video Description}

\author{$^\star$Xuyang Shen$^{2}$, $^\star$Dong Li$^{1}$, $^\star$Jinxing Zhou$^{3}$, Zhen Qin$^{2}$, Bowen He$^{2}$, Xiaodong Han$^{2}$, Aixuan Li$^{4}$, \\
        Yuchao Dai$^{4}$, Lingpeng Kong$^{5}$, Meng Wang$^{3}$, Yu Qiao$^{1}$, $^\textrm{\Letter}$ Yiran Zhong$^{1}$
\vspace{2mm} \\
$^{1}$Shanghai Artificial Intelligence Laboratory, 
$^{2}$OpenNLPLab,
$^{3}$Hefei University of Technology, \\
$^{4}$Northwestern Polytechnical University,
$^{5}$The University of Hong Kong 
}
\maketitle

\blfootnote{$^{\star}$These authors have equal contributions. $^\textrm{\Letter}$Yiran Zhong is the corresponding author (e-mail: zhongyiran@gmail.com).}

\begin{abstract}
\vspace{-2mm}
We explore a new task for audio-visual-language modeling called fine-grained audible video description (FAVD). It aims to provide detailed textual descriptions for the given audible videos, including the appearance and spatial locations of each object, the actions of moving objects, and the sounds in videos. Existing visual-language modeling tasks often concentrate on visual cues in videos while undervaluing the language and audio modalities. On the other hand, FAVD requires not only audio-visual-language modeling skills but also paragraph-level language generation abilities. We construct the first fine-grained audible video description benchmark (FAVDBench) to facilitate this research. For each video clip, we first provide a one-sentence summary of the video, \emph{\ie}, the caption, followed by 4-6 sentences describing the visual details and 1-2 audio-related descriptions at the end. The descriptions are provided in both English and Chinese. We create two new metrics for this task: an \emph{EntityScore} to gauge the completeness of entities in the visual descriptions, and an \emph{AudioScore} to assess the audio descriptions. As a preliminary approach to this task, we propose an audio-visual-language transformer that extends existing video captioning model with an additional audio branch. We combine the masked language modeling and auto-regressive language modeling losses to optimize our model so that it can produce paragraph-level descriptions. We illustrate the efficiency of our model in audio-visual-language modeling by evaluating it against the proposed benchmark using both conventional captioning metrics and our proposed metrics. We further put our benchmark to the test in video generation models, demonstrating that employing fine-grained video descriptions can create more intricate videos than using captions. Code and dataset are available at {\color{red}{\textit{\href{https://github.com/OpenNLPLab/FAVDBench}{https://github.com/OpenNLPLab/FAVDBench}}}}. Our online benchmark is available at \color{red}{\textit{\href{http://www.avlbench.opennlplab.cn}{www.avlbench.opennlplab.cn}}}. 

\newpage

\end{abstract}

\vspace{-6.3mm}
\input{1_introduction}

    \input{table1}

\input{2_related_work}

    \begin{figure*}[t]
      \centering
      \includegraphics[width=0.88\textwidth]{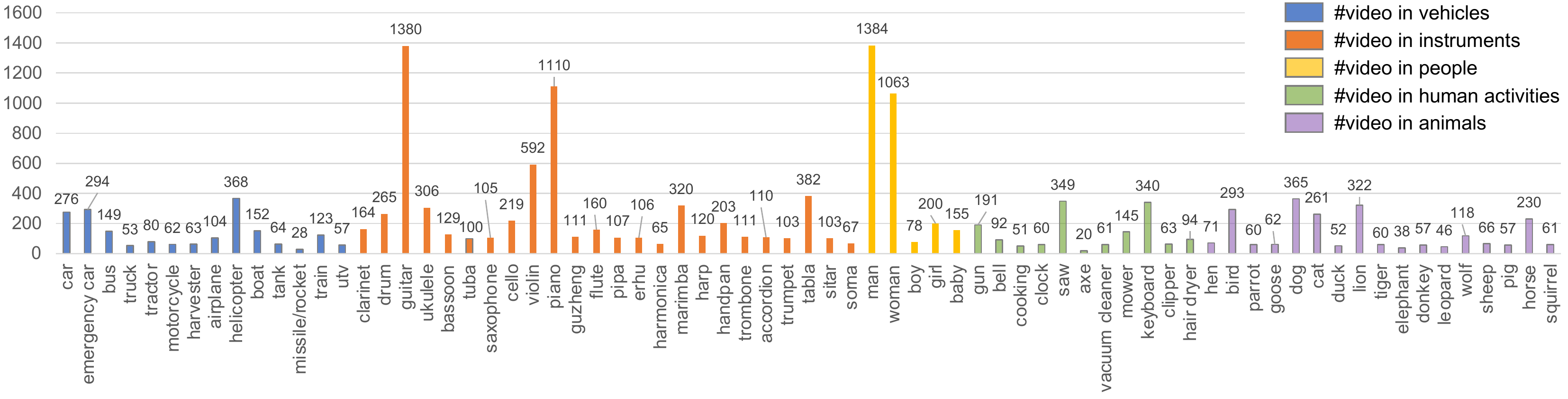}
      \vspace{-4mm}
      \caption{\textbf{Video distribution of 5 major categories and 71 sub-classes in FAVDBench.} The value represents the total occurrence in visual of each class. The vehicles class is colored blue, contains 14 sub-classes. There are 24 sub-classes in instruments, colored orange. The people category includes man, woman, boy, girl, and baby, which are colored yellow. The category of human-related activities is colored green and contains 11 sub-classes. There are 17 sub-classes in animals, colored purple. Best view in color.}
      \vspace{-5mm}
      \label{fig:dataset_stat}
    \end{figure*}

\input{3_video_description}

\input{4_methodology}

\input{5_experiments}

\input{6_conclusion}

\input{7_appendix}

\clearpage

{\small
\bibliographystyle{ieee_fullname}
\bibliography{egbib}
}

\end{document}

%% file: 1_introduction.tex
\vspace{-6mm}
\section{Introduction}
\vspace{-1mm}
\label{sec:intro}

\begin{figure*}[t]
  \centering
   \includegraphics[width=0.88\textwidth]{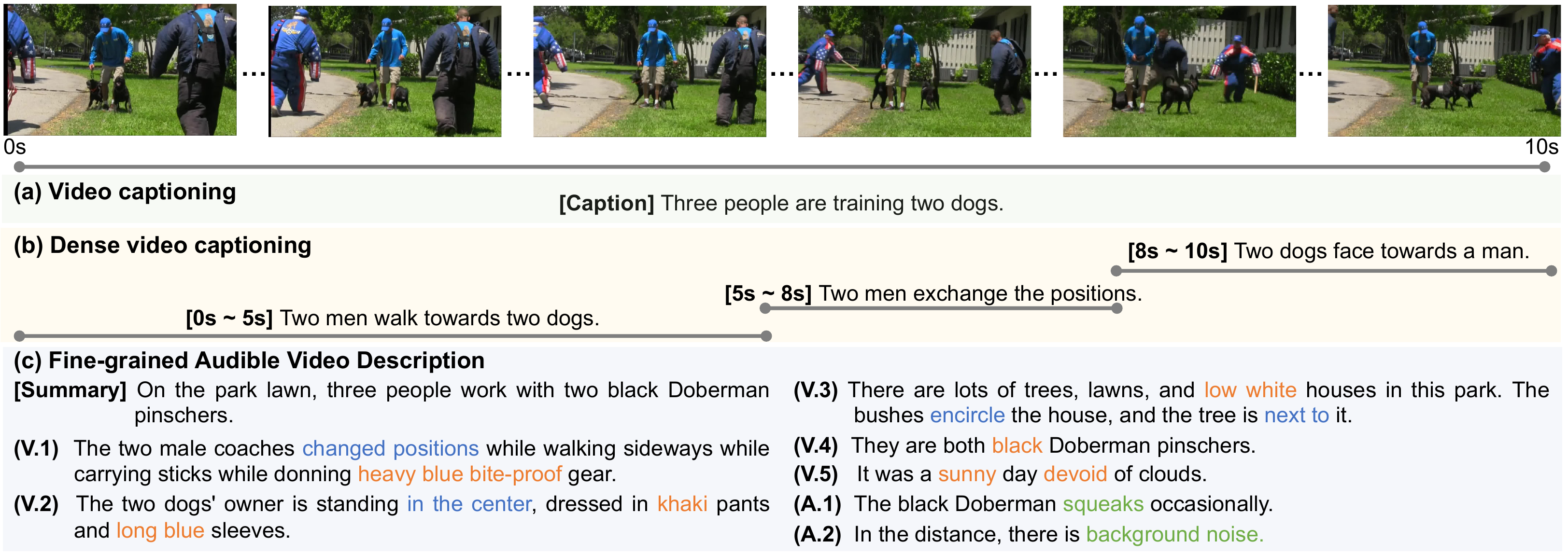}
   \vspace{-4mm}
   \caption{\textbf{Comparison of the proposed FAVD task with existing captioning tasks.} (a) Video captioning (VC) uses one sentence to describe the main content of the video. (b) Dense video captioning aims to localize the multiple temporal events and generate corresponding descriptions. Both VC and DVC describe the salient events in videos while losing many details, such as the appearance of objects, spatial relations, and sounds. (c) The proposed FAVD tries to generate a paragraph-level description that contains the caption, named as \textbf{Summary}, and the audio-visual descriptions, abbreviated as \textbf{A.} and \textbf{V.}.}
   \vspace{-5mm}
   \label{fig:dataset_intro}
\end{figure*}

Language serves as the primary form of human communication, providing not only complementary information to other modalities such as vision and audio, but also an efficient means of exchanging information~\cite{leech1974semantics,origgi2000evolution,hayakawa2019language}. For example, we can use voice navigation to guide us to our destination. Visually impaired people can watch a movie by listening to its narration. The former shows that the language can provide complementary information to other modalities, while the latter indicates that the language can carry the most information in other modalities.

Recent multi-modal modeling tasks attempt to connect the language to other modalities, including image/video captioning~\cite{Vinyals_2015_CVPR,ordonez2011im2text,li2022comprehending,gurari2020captioning,pan2020x,pei2019memory,wang2018reconstruction,zhang2020object,song2022memorial,song2022contextual,zhou2021semi}, text-to-image/video synthesis~\cite{huang2022dse,wu2022nuwa,gu2022vector,ruan2021dae,ramesh2021zero,ramesh2022hierarchical,hong2022cogvideo,singer2022make,ho2022imagen,villegas2022phenaki}, text-driven image/video manipulation~\cite{nam2018text,li2020manigan,bar2022text2live,zhou2022audio,zhou2023audio}, and \etc.
However, in these tasks, since the language is frequently used to provide complementary information to other modalities, they often fail to give a fine-grained description of the exchange of information between modalities, and only consider the simplified language, \ie, one-sentence captions. 
Due to the conciseness of captions, only salient objects and activities have got descriptions. As a result, the information carried by the captions will be much less than that carried by other modalities, causing significant information loss when exchanging information from other modalities to language. 

In this paper, we consider language as a tool for conveying information from other modalities in multi-modal modeling and propose a new task called Fine-grained Audible Video Description (FAVD).
We compare it with existing captioning tasks in Fig.~\ref{fig:dataset_intro}. Video captioning aims to generate one concise sentence to summarize the whole video. Dense video captioning first detects the multiple temporal events and then provides one caption for each event. The generated sentences of these two tasks merely cover the main objects or movements while losing many details when translating the video information into language.
Unlike this, the FAVD task requires a model to describe videos in a similar way that humans do, \ie, starting with an overview of the video and then focusing on each fine-grained detail to concrete the description. In this case, most video information can be preserved in the language modality. Since a video contains both visual and audio signals, we also include audio descriptions in our task. 

FAVD is a non-trivial task. In addition to long-text modeling abilities, it calls for a finer-grained visual understanding than earlier video captioning tasks, \ie, it needs to recognize all objects and actions in videos, as well as describe the appearance and spatial locations of each object and the sounds in videos. To facilitate this task, we construct the first fine-grained audible video description benchmark (FAVDBench), which allows the model to be trained in a supervised manner.

FAVDBench is made up of over 11,000 video clips culled from millions of YouTube videos and each clip is sourced by querying more than 70 life-related categories to fulfill the diversity of the benchmark. We annotate the video descriptions based on human behaviors. The annotations of each video clip begin with a one-sentence caption that describes the salient objects and activities, followed by 4-6 sentences that describe visual details including the appearance and spatial locations of each object, and the actions or movements of moving objects. To make the task consider audio information, we include 1-2 sentences of audio-related descriptions at the end of the whole descriptions. The descriptions are provided in both English and Chinese with human translation. An annotation example can be found in Fig.~\ref{fig:dataset_intro} and Fig.~\ref{fig:annotation}.
We design two new metrics for the FAVD task. One called \emph{EntityScore}, which assesses the completeness of the information that is transferred from the videos to the descriptions by gauging the completeness of entities in the visual descriptions. The other is \emph{AudioScore}, which measures the audio description in the feature space of a pretrained audio-visual-language model.

We provide a baseline model for this new task. The model is based on an existing end-to-end video captioning model~\cite{lin2022swinbert} with an additional audio branch. We also extend the visual-language transformer to the audio-visual-language transformer in order to mix multi-modal tokens. Existing video captioning models often adopt Masked Language Modelling (MLM) loss to optimize the caption generation. However, because our task requires the model to describe fine-grained details, the model should also be capable of general language modeling. To this end, we include the auto-regressive language modeling (ALM) loss in our baseline model.

We extensively evaluate the baseline model against the proposed FAVDBench using both the conventional captioning metrics and our proposed metrics and demonstrate the effectiveness of our model in audio-visual-language modeling and fine-grained description generation. To qualitatively illustrate the information preservation in modality information exchange, we further put our benchmark to the test in video generation models, showing more intricate videos than using captions.

%% file: table1.tex
\begin{table*}[t]\small
\caption{\textbf{Statistics of FAVDBench and other existing video captioning datasets.} 
FAVDBench is the first dataset to provide audible descriptions of videos. ``\#Sentence'' and ``\#Word'' denote the average number of sentences and words contained in the annotation per video, respectively. We also display the part-of-speech tagging (POS tag) in the percentage of each dataset. FAVDBench has the highest percentage of adjective words (Adj.) which indicates the rich annotation of fine-grained audio-visual details.}
\centering
\label{tab:stat_comparsion}
\vspace{-2mm}
 \begin{tabular}{p{2.2cm}<{\centering}p{1.cm}<{\centering}p{1.2cm}<{\centering}p{1.cm}<{\centering}p{1.4cm}<{\centering}p{1.4cm}<{\centering}p{1.4cm}<{\centering}p{1.2cm}<{\centering}p{1.2cm}<{\centering}p{1.2cm}<{\centering}}
  
\toprule[0.8pt]\noalign{\smallskip}
\multirow{2}{*}{Dataset}  & \multicolumn{2}{c}{Video}       & \multicolumn{4}{c}{Annotation}                    & \multicolumn{3}{c}{POS tag} \\
\noalign{\smallskip}
\cmidrule(r){2-3} \cmidrule(r){4-7} \cmidrule(r){8-10}
                                & \#Clip  & DUR. (h)    & Audio       & \#Sentence  & \#Word    & Vocabulary & \%Adj.          & \%Noun         & \%Prep.       \\ \midrule
MSVD~\cite{chen2011collecting} & 1,970   & 5.3          & \ding{56}    & 35.5       & 308.3      & 13,010     & 2.6           & 31.8           & 7.7           \\
MSR-VTT~\cite{xu2016msr}       & 10,000  & 41.2         & \ding{56}    & 20.0       & 185.7      & 29,316     & 4.8           & 33.9           & 11.5          \\
VATEX~\cite{wang2019vatex}     & 41,250  & 114.6        & \ding{56}    & 20.0       & 291.8      & 82,654     & 4.4           & 31.8           & 12.4           \\
TVC~\cite{lei2020tvr}          & 21,793  & 461.3        & \ding{56}    & 5.0        & 67.0       & 57,100     & 2.2           & 36.4           & \textbf{12.7}  \\
YouCook\uppercase\expandafter{\romannumeral2}~\cite{ZhXuCoAAAI18} & 15,433  & 176.0        & \ding{56}    & 1.0        & 7.9       & 2,583      & 4.1           & \textbf{40.3} & 11.6            \\ \midrule
FAVDBench                     & 11,424  & 24.4         & \ding{52}    & 12.6       & 218.9       & 73,245     & \textbf{13.0}    & 30.5        & 12.4            \\

\toprule[0.8pt]
\end{tabular}
\vspace{-7mm}
\end{table*}

%% file: 2_related_work.tex
\section{Related Work}
\label{sec:related_work}

\noindent
\textbf{Video captioning.}
\label{subsec:video_caption}
Video captioning aims to describe the main content in a video with one concise natural language sentence.
The pioneer works mainly follow the sequence to sequence pipeline, where the Convolutional Neural Network is used to encode the video frame features and the Recurrent Neural Network is used to decode the predicted sentence~\cite{venugopalan2015sequence,xu2017learning,wang2018reconstruction,pei2019memory}.
Some following works adopt the Transformer architecture for efficient captioning~\cite{fang2020video2commonsense,luo2020univl,lin2022swinbert,lu2022linear,qin2022devil,lu2022linear,cheng2022implicit,zhong20183d,zhang2021depth}.
Recently, Tang \etal~\cite{tang2021clip4caption} use the pretrained CLIP~\cite{radford2021learning} to extract advanced text and visual features that improve the captioning performance significantly.
Besides, there are several works that try to explore the spatio-temporal clues~\cite{aafaq2019spatio} or object relations~\cite{zhang2020object,pan2020spatio} to give more accurate descriptions.
However, it is hard for these captioning models to detail the rich semantics in a video with only one sentence. 
Our proposed FAVD task carries the FAVDBench dataset which provides fine-grained annotations allowing us to train an intelligent robot that can tell rich video details.

\noindent
\textbf{Dense video captioning.}
\label{subsec:dense_video_caption}
Since there are usually complex content and various events contained in a long and untrimmed video, it is insufficient to describe the video with only one sentence.
For this reason, dense video captioning is proposed ~\cite{krishna2017dense} which aims to automatically localize the multiple temporal events and generate corresponding captions.
Most of the related works follow the ``localize then describe'' scheme~\cite{zhou2018end,li2018jointly,mun2019streamlined,wang2018bidirectional,sun2021getam,sun2023munet} which first predicts many event proposals and then generates its captions.
Unlike this, Wang \etal~\cite{wang2021end} propose an end-to-end captioning model that decodes the event proposals and captions parallelly. 
Some methods utilize the multi-modal information such as audio~\cite{iashin2020multi,iashin2020better,rahman2019watch,zhou2023improving} and motion~\cite{chen2020learning} in videos for better captioning.
Although these methods achieve satisfactory performance on dense video captioning, they all manipulate the frame-level features that prohibit the network from learning fine-grained details, which is indeed the essential expectation of the proposed FAVD task.
Unlike them, the proposed baseline framework encodes the patch-level video representations with the aggregation of multi-modal tokens that enables to give more detailed descriptions.

\noindent
\textbf{Audio captioning.}
\label{subsec:audio_caption}
Unlike video captioning and dense video captioning which mainly focus on the visual domain, audio captioning aims to directly generate text description for audio in the wild~\cite{kim2019audiocaps,mei2021encoder,eren2021audio,liu2021cl4ac,ye2021improving}.
In the initial work, Kim \etal~\cite{kim2019audiocaps} propose an encoder-decoder method, where the Bi-LSTM is used to generate the captions word by word.
Recently, Mei \etal~\cite{mei2021audio} use the Transformer backbone to encode the long-range dependencies in the audio signal and output the whole predicted sentence.
Though these methods enable us to describe the audio to some extent, we argue that it is hard for accurate audio captioning without giving any visual information.
FAVD takes both audio and visual signals as inputs, which eases the description generation.

\noindent
\textbf{Audio-visual-language dataset.}
Existing captioning-related multi-modal datasets mainly focus on two modalities: the \emph{vision-language} datasets for (dense) video captioning, such as MSVD~\cite{chen2011collecting}, MSR-VTT~\cite{xu2016msr}, VATEX~\cite{wang2019vatex}, TVC~\cite{lei2020tvr}, and YouCook\uppercase\expandafter{\romannumeral2}~\cite{ZhXuCoAAAI18};
or the \emph{audio-language} datasets for audio captioning, such as AudioCaps~\cite{kim2019audiocaps} and Clotho~\cite{drossos2020clotho}.
A few video captioning works try to introduce the audio track contained in videos of vision-language datasets, but the audio actually limits to speech domain~\cite{rohrbach2015dataset,ZhXuCoAAAI18,monfort2021spoken,lei2020tvr}.
While videos in the proposed FAVDBench are collected from various life-related scenes that extend the diversity of the audio. Besides, the audio and visual signals are usually semantic corresponding which makes it a truly audio-visual-language dataset.

%% file: 3_video_description.tex
\vspace{-2mm}
\section{The FAVDBench}
\vspace{-1mm}
\label{sec:dataset}

\subsection{Task description}
\vspace{-1mm}
The FAVD task involves generating fine-grained textual descriptions for the given audible videos that include information about appearance and spatial location of each object, the movements of moving objects, and the sounds in the videos. 
Specifically, given an audible video clip, this task aims to generate a paragraph with 6-9 sentences. The first sentence needs to be an overview of the video clip that emphasizes the key visual elements as well as the main event. The next 4-6 sentences should then be used to describe the minute visual details of each object, \ie, the rich semantics that each one possesses and the function that it serves in the video. The final 1-2 sentences characterize the audio information contained in the video clip, describing what is producing the sound or what it sounds like. We compare the annotation differences between the FAVD and previous video captioning tasks in Fig.~\ref{fig:dataset_intro}.

\vspace{-1mm}
\subsection{Dataset statistics}\label{subsec:statistics}
\vspace{-1mm}
We construct a benchmark called FAVDBench to facilitate the FAVD task. We adopt the technique described in VGGSound~\cite{arandjelovic2018objects} to make sure the audio and visual clips match the intended semantics. All videos are collected from YouTube under \emph{Creative Commons} license. We take into account the number of sounding objects when collecting video clips as we find the difficulty of audio description is positively correlated with the number of sounding objects. As a result, 60.2\% of the video clips have one sounding object and the rest have 2-3 sounding objects.

The FAVDBench consists of 11,424 audible video clips and 143,548 sentences of annotations. Each video clip has a description of 12.6 sentences, 218.9 word tokens on average. Among the clips, 5,289 videos are trimmed to 5 seconds and 6,135 videos to 10 seconds. The total video duration is 24.4 hours and 2.6 million frames. To avoid data leakage and maintain diversity, we enforce a minimum interval of 5 seconds between adjacent videos. We also manually filter clips with similar content. The benchmark is divided into 4 partitions: 7,500 for training, 1,500 for validation, 1,000 for testing, and 1,424 withhold for online benchmarking. The withhold partition will only be visible for contestants in the future FAVD Benchmark competition. 

We compare FAVDBench with other video captioning datasets in Table~\ref{tab:stat_comparsion}. In terms of context differences, we include 5 major categories for video content, \ie, vehicles, instruments, animals, people, and common human activities. We further divide the 5 major categories into 71 sub-classes and the distribution of each category/subclass is plotted in Fig.~\ref{fig:dataset_stat}. By contrast, the contexts of existing captioning datasets TVC~\cite{lei2020tvr} and YouCook\uppercase\expandafter{\romannumeral2}~\cite{ZhXuCoAAAI18} are restricted to movie and cooking scenarios while the MSVD~\cite{chen2011collecting}, MSR-VTT~\cite{xu2016msr}, and VATEX~\cite{wang2019vatex} merely include human activities. Besides, we also provide audio descriptions for each video clip in FAVDBench, which distinguishes us from other datasets. Furthermore, since our benchmark provides fine-grained descriptions for video objects, the percentage of adjectives in our dataset is more than 3 times higher than in other datasets.

\subsection{Annotation details}\label{subsec:annotation}
\begin{figure}[t]
  \centering
  \vspace{0mm}
  \includegraphics[width=0.48\textwidth]{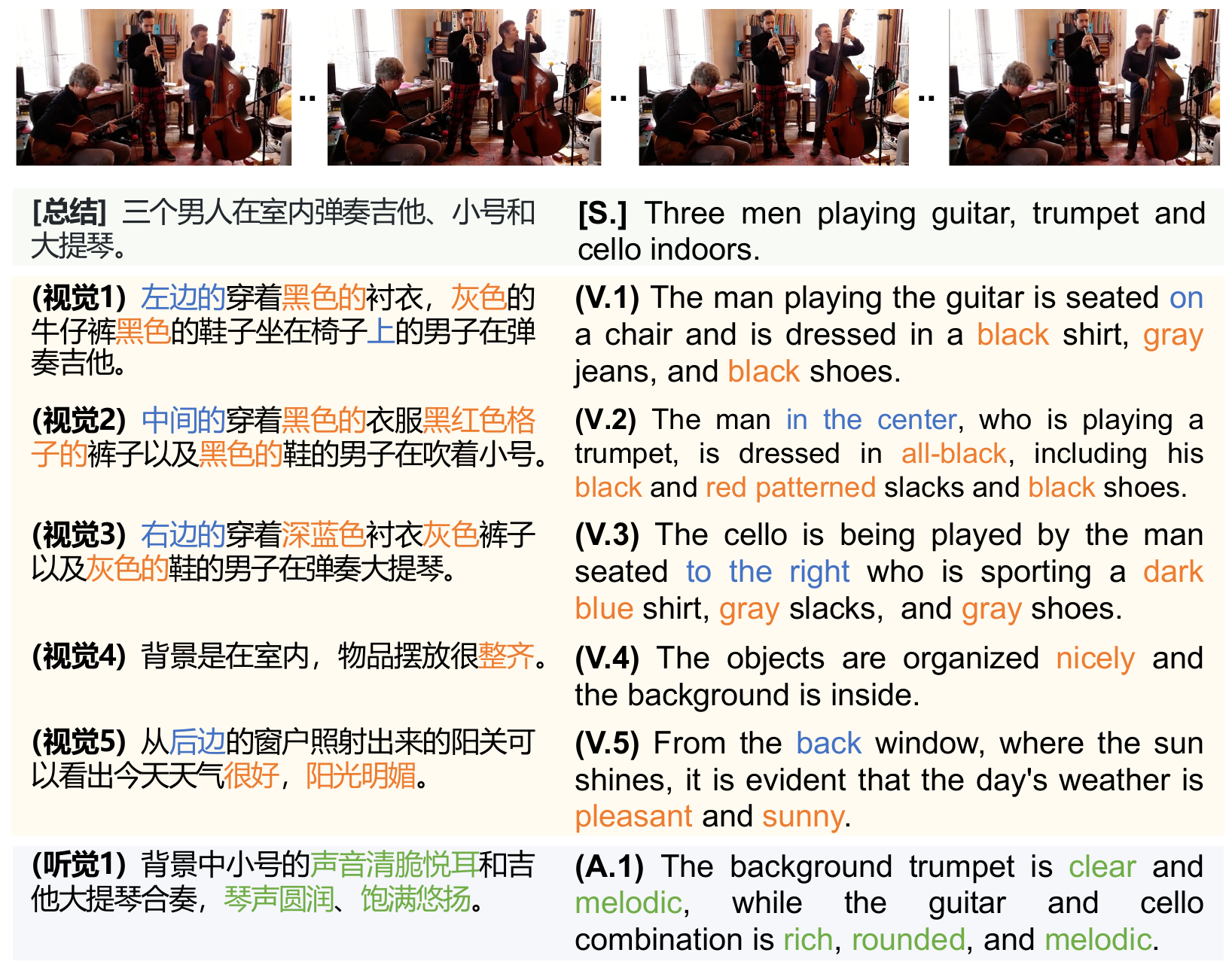}
  \vspace{-7mm}
  \caption{\textbf{An example of FAVDBench annotation.} For each video clip, we provide both Chinese and English annotations, each contains 1 summary \textbf{[S.]}, 4-6 detailed visual descriptions \textbf{[V.]} and 1-2 audio-related descriptions \textbf{[A.]}. Details of spatial locations, appearance, and sounds are highlighted in blue, orange, and green.}
  \vspace{-6mm}
  \label{fig:annotation}
\end{figure}

\begin{figure*}[ht]
  \centering
  \vspace{-3mm}
  \includegraphics[width=0.88\textwidth]{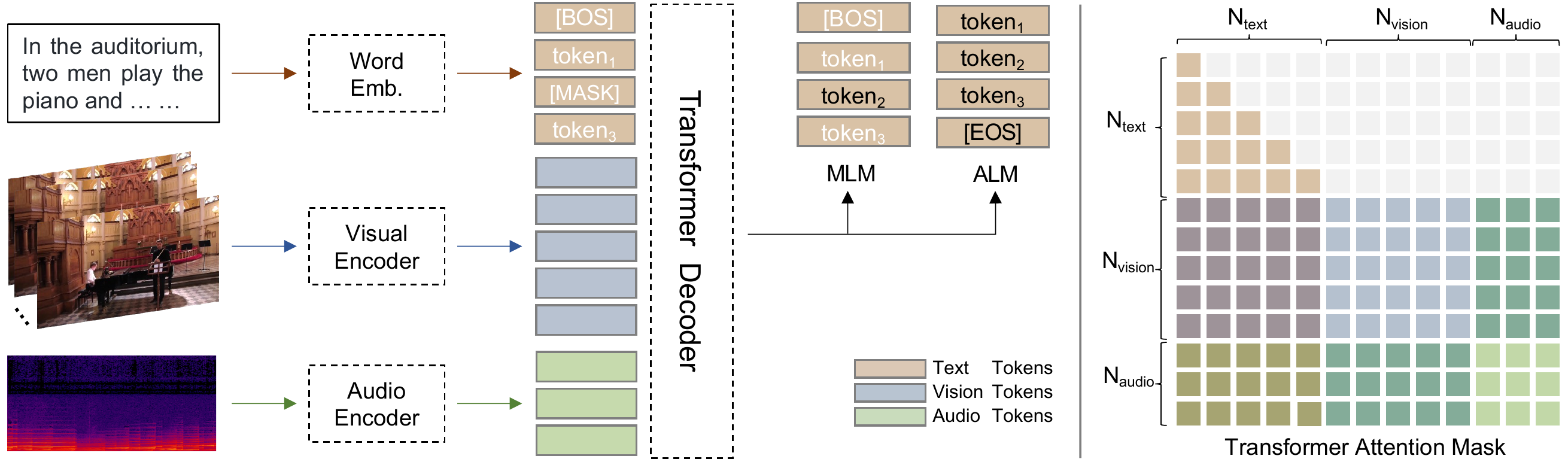}
  \vspace{-4mm}
    \caption{\textbf{Overview of AVLFormer.} It consists of a word embedding, a visual encoder, an audio encoder, and a transformer decoder. We adopt the video swin transformer and patchout audio transformer as the visual encoder and audio encoder, respectively. They extract visual and audio features from video frames and audio. Masked language modeling and auto-regressive language modeling are configured in training. The attention mask strategy of AVLFormer is illustrated on the right, where the masked attention is colored in gray. The tokens and attention masks of text, vision, and audio are colored brown, blue, and green, respectively.}
  \vspace{-5mm}
  \label{fig:model}
\end{figure*}

FAVDBench provides both English and Chinese annotations. All video clips are primarily annotated in Chinese using crowdsourcing and then translated to English with human translation. To collect large-scale fine-grained descriptions of videos, we design several rules of annotation. Specifically, each video is required to be watched completely, and then described by 1 summary sentence, at least 4 visual-related points, and at least 1 audio-related point. The summary only covers high-level information about the most salient event that the video tries to convey. The order to describe the visual and audio details follows from the most salient objects to the background. The statuses of objects are required to be thoroughly captured, including the actions, characteristics (adjective), and relative spatial relations (preposition). Furthermore, each sentence is required to be larger than 5 Chinese characters. Any subjective thoughts and speculative words (e.g., ``maybe'') are forbidden to appear in labels, especially in the audio description. In addition, information beyond the video and audio is not allowed to include in the annotations, such as the wiki of objects, subtitles, and speech contents. 
A video sample with bilingual annotations is shown in Fig.~\ref{fig:annotation}.

To reduce the annotation bias and ensure the quality of annotation, the annotators are restricted to around 60 people, and all of them are native speakers of Chinese. Each video is annotated by one worker, checked by 2 peer workers, and examined by grammar-checking tools. The English and the Chinese version of descriptions are congruent to prevent any changes to the semantics of sentences.

\input{table2}

\subsection{Evaluation metrics}\label{subsec:eval_metrics}
The generated descriptions should be evaluated in terms of semantics rather than word-for-word precision. For example, we would like to know whether the entities in the videos are addressed and the sounding objects are properly described, rather than whether they are using the same way to describe these things. Existing video caption metrics, such as BLEU~\cite{papineni-etal-2002-bleu}, ROUGE~\cite{10.3115/1073445.1073465}, Meteor~\cite{denkowski-lavie-2014-meteor}, and CIDEr ~\cite{7299087} often concentrate on word-for-word precision by measuring the token similarity between the generated and ground truth texts, which does not meet our needs.
Therefore, to properly evaluate the generated descriptions, we design two new metrics called EntityScore $\mathbb{ES}$ and AudioScore $\mathbb{AS}$.

\emph{EntityScore} measures the extent to which consistently referred words or series of words, known as entities and often manifested as nouns, in the predicted text match those in the annotated text. We use an off-the-shelf Natural Language Toolkit library to extract nouns as entities. The mathematical expression of the EntityScore $\mathbb{ES}$ is as follows:
\begin{equation}
\small
     \begin{array}{lll}
     \vspace{1ex}
      R(\mathbf p,\mathbf r)=\frac{\#\{ \mathbf p \cap \mathbf r \}}{\#\{\mathbf r\}}, 
     C(\mathbf p,\mathbf r)=\frac{\cos(\mathrm{T5}(\mathbf p), \mathrm{T5}(\mathbf r))+1} 2, \\
     \mathbb{ES}(\mathbf p,\mathbf r)= \frac{2 R(\mathbf p,\mathbf r) C(\mathbf p, \mathbf r)}{R(\mathbf p,\mathbf r) + C(\mathbf p, \mathbf r)}, 
     \end{array}
\end{equation} 
where $\mathbf{p}\in \mathbb R^{n}$ is the predicted entities from the model predictions, and $\mathbf{r}\in \mathbb R^{m}$ is the ground truth entities, $n$ and $m$ are the entity numbers. 

$\text{\#}\{ \mathbf r \}$ means the number of entities in the reference, and $\text{\#}\{ \mathbf p \cap \mathbf r \}$ counts the number of correctly described entities in the prediction. ${\cos}$ denotes the cosine similarity between two vectors and T5 represents the pretrained model T5~\cite{2020t5} used to extract entity features. Besides, it is normalized into the range [0,1] to guarantee its positive. Therefore, ${R(\mathbf p,\mathbf r)}$ indicates the recall of the predicted entity that is described using the same word as the reference and ${C(\mathbf p,\mathbf r)}$ measures the comprehensiveness of semantics between the predicted entities and the ground truth entities. In other words, $R$ focuses on syntactic level accuracy on entities while $C$ concentrates on semantic level entity similarities. The $\mathbb{ES}$ considers both syntactic and semantic level accuracy. However, $\mathbb{ES}$ has a limitation in that it places emphasis solely on the presence of entities while disregarding their placement and frequency.

\emph{AudioScore} assesses the accuracy of audio descriptions by computing the product of the extracted audio-visual-text unit features,
where CLIP~\cite{radford2021learning} is used to extract features for video frames and the corresponding descriptions and PaSST~\cite{koutini2021efficient} is used for audio waves. Moreover, we fine-tune these models using contrastive learning on FAVDBench to ensure that the latent features are mapped to a shared space. Specifically, the AudioScore $\mathbb{AS}$ is defined as:
\begin{equation} 
\small
    \begin{gathered}
      \mathbf e_a=\mathrm{PaSST}(\mathbf A),\mathbf e_v=\mathrm{CLIP}(\mathbf V),  \mathbf e_t = \mathrm{CLIP} (\mathbf T), \\
    s = \left(\frac{1}{2} \cos(\mathbf e_a, \mathbf e_t)  + \frac{1}{2} \cos(\mathbf e_a, \mathbf e_v) + 1 \right) \times 0.5,\\
    \mathbb{AS}(\mathbf A,\mathbf V,\mathbf T) = \mathbf f(s),
    \mathbf f(x) = a \exp (-b\exp (-c x)), 
    \end{gathered}
\label{eq:audio_score} 
\end{equation}
wher $\mathbf{A}$ and $\mathbf{V}$ denote the audio and visual frames of one video, respectively, while $\mathbf{T}$ is the predicted audio description. $\mathbf{cos}$ and  $\mathbf{f}$ denote cosine similarity and a specific form of Gompertz function, respectively. We set c=10 empirically, and choose values for a and b ($a=\frac{1}{e^{-0.69e^{-10}}}$, b=0.693) to force specific values of x and f(x) (x=1, f(x)=1 for $a=\frac{1}{e^{-0.69e^{-10}}}$, and x=0, f(x)=0.5 for b=0.693).
We select the last one or two sentences from the whole generated paragraph description as $\mathbf{T}$, as we find that the model prefers describing audio information at the end of the description, mimicking the ground truth. 
Therefore, we report the Top-1 and Top-2 accuracy as the $\mathbb{AS}$ score for the last one or two sentences. 
We use the Top-1 score by default.

%% file: table2.tex
\begin{table*}[t]\small
\caption{\textbf{Comparison with different methods on FAVDBench.} We report different backbone settings for PDVC, SwinBERT, BMT and AVLFormer. In the backbone freeze (FRZ.) column, \textbf{-} represents that input is not available; \ding{220} represents that is trainable; \ding{91} represents frozen. The visual and audio backbone of BMT is offline by default. For all metrics, higher values are better. The maximum human evaluation score of ten (10) is regarded as representing the ground truth.}
\vspace{-3mm}
\centering
\label{tab:favd_performance}
 \begin{tabular}{p{2.4cm}<{\centering}p{0.7cm}<{\centering}p{0.7cm}<{\centering}p{0.8cm}<{\centering}p{0.8cm}<{\centering}p{0.8cm}<{\centering}p{0.8cm}<{\centering}p{1.2cm}<{\centering}p{1.3cm}<{\centering}p{1.3cm}<{\centering}p{1.3cm}<{\centering}}
\toprule[0.8pt]
\multirow{2}{*}{Method} & \multicolumn{2}{c}{Backbone FRZ.} & \multicolumn{5}{c}{Conventional Metric}  & \multicolumn{2}{c}{Proposed Metric} & \multirow{2}{*}{\shortstack{Human\\Evaluation}}\\ \cmidrule(r){2-3} \cmidrule(r){4-8} \cmidrule(r){9-10} 
                                & Visual            & Audio         & B@1            & B@4            & Meteor          & CIDEr          & Clipscore     & EntityScore     & AudioScore   \\ \midrule
PDVC~\cite{wang2021end}         & \ding{220}        & \textbf{-}    & 34.91          & 6.33           & 13.80           & 6.27           & 65.77          & 33.45          & 56.88            & 3.7   \\ 
PDVC w. Audio                   & \ding{220}        & \ding{220}    & 35.59          & 6.47           & 14.49           & 13.13          & 66.15          & 34.12          & 60.72            & 4.1   \\ \midrule
SwinBERT~\cite{lin2022swinbert} & \ding{220}        & \textbf{-}    & 39.68          & 9.05           & 16.78           & 21.69          & 68.13          & 39.60          & 62.60            & 6.0   \\
SwinBERT~\cite{lin2022swinbert} & \ding{91}         & \textbf{-}    & 41.41          & 9.08           & 17.17           & 23.03          & 67.98          & 41.00          & 58.72            & 5.9   \\ \midrule
BMT~\cite{iashin2020better}     & \ding{91}         & \ding{91}     & 41.28          & 9.23           & 16.57           & 16.46          & 65.00          & 35.99          & 61.03            & 6.4    \\ \midrule
AVLFormer                       & \ding{220}        & \ding{220}    & \textbf{44.10} & \textbf{10.29} & \textbf{18.36}  & \textbf{29.85} & 69.74          & \textbf{44.46} & \textbf{63.88}   & \textbf{8.2}   \\
AVLFormer                       & \ding{91}         & \ding{220}    & \textbf{44.10} & 10.23          & 18.24           & 26.31          & 68.98          & 43.47          & 61.53            & 7.9   \\
AVLFormer                       & \ding{220}        & \ding{91}     & 42.82          & 10.16          & 17.96           & 25.45          & \textbf{70.07} & 43.05          & 62.84            & 7.8   \\
AVLFormer                       & \ding{91}         & \ding{91}     & 42.35          & 9.84           & 17.73           & 25.65          & 69.33          & 43.13          & 61.90            & 7.6   \\
\toprule[0.8pt]
\end{tabular}
\vspace{-6mm}
\end{table*}

%% file: 4_methodology.tex
\section{The AVLFormer}
\label{sec:method}

\vspace{-1.5mm}
We propose a transformer-based model called Audio-Visual-Language Transformer (AVLFormer) as a baseline method for the FAVD task. AVLFormer is in a form of encoder-decoder structures as shown in Fig.~\ref{fig:model}. A visual encoder and an audio encoder are adapted to process the video clips and audio, respectively. Similar to previous video captioning models~\cite{lin2022swinbert}, AVLFormer also takes textual descriptions as input and uses a word tokenizer and a linear embedding to embed the text. The output of AVLFormer is the fine-grained descriptions of the input videos.

\noindent
\textbf{The encoders.}\label{subsec:econder} 
We employ the video swin transformer~\cite{liu2022video} as our visual encoder as it achieves state-of-the-art performance in video understanding tasks~\cite{liu2022video, lin2022swinbert}. The weights of the visual encoder are pretrained on the ImageNet-22k~\cite{deng2009imagenet} and the Kinetics-600~\cite{carreira2018short} datasets. The visual features $\mathbf F_{v} \in \mathbb{R}^{N_{v} \times d_v}$, where $d_{v}=512$ is the feature dimension and $N_{v}=784$ is the patch sequence length. We adopt the patchout audio transformer (PaSST)~\cite{koutini2021efficient} as our audio encoder. We pretrain it on the ImageNet-1K~\cite{deng2009imagenet} and the AudioSet~\cite{gemmeke2017audio} datasets. The audios are firstly processed by a Fast Fourier Transform and a Mel feature extractor and then sent to PaSST. The audio features $\mathbf F_{a} \in \mathbb{R}^{N_{a} \times d_{a}}$, where $d_{a}=768$ is the dimension and $N_{a}=473$ is the sequence length. For the text input, we apply a word tokenizer and a linear embedding to embed the text as $\mathbf F_{t} \in \mathbb{R}^{N_{t} \times d_{t}}$, where $d_{t}=768$ and $N_{t}=300$.

\noindent
\textbf{Audio-visual-language transformer.}\label{subsec:AVLTransformer} 
Before passing the audio features $\mathbf F_{a}$, visual features $\mathbf F_{v}$, and text embedding $\mathbf F_{t}$ to the audio-visual-language transformer (AVLFormer), a linear projection layer is employed to unify their feature dimensions to $d=768$. AVLFormer takes the concatenated feature $\mathbf F_{avt} \in \mathbb{R}^{(N_t + N_v + N_a) \times d}$ of $\mathbf F_{a}$, $\mathbf F_{v}$, and $\mathbf F_{t}$ as input and processes it with a stack of multi-head self-attention modules. The number of layers of the transformer is set to 12. We illustrate the masking strategy of AVLFormer in Fig.~\ref{fig:model}. Specifically, we use the auto-regressive language model masking strategy to mask the text attention matrix, which ensures that the predicted tokens are only dependent on the previous tokens. For the audio and visual tokens, since there is no ordering restriction for these modalities, we use full attention matrices instead. We also make audio and visual tokens visible to text tokens, ensuring that other modality information can contribute to text generation.

\noindent
\textbf{Loss functions.}\label{subsec:loss} 
Previous video captioning methods~\cite{lei2021less, lin2022swinbert} mainly adopt the masked language modeling loss to make the model concentrate on salient objects and actions, whereas previous vision-language generation methods~\cite{li2022blip} often use auto-regressive language modeling loss for longer text generation. For the FAVD task, since it requires not only fine-grained descriptions for each object but also paragraph-level text generation capability, we employ both the masked language modeling loss $\mathcal{L}_{\mathrm{MLM}}$ and the auto-regressive language modeling loss $\mathcal{L}_{\mathrm{ALM}}$ to train our AVLFormer. In masked language modeling, AVLFormer is required to recover the words that are replaced by [\texttt{MASK}] symbol, as shown in Fig.~\ref{fig:model}. In this way, the model is forced to recuperate the information based on audio-visual information. In auto-regressive language modeling, AVLFormer is asked to understand the semantic context for the entire sentence. Note that AVLFormer works in an auto-regressive manner in inference, the $\mathcal{L}_{\mathrm{ALM}}$ can also mitigate the gap between the training and the inference. The total object function $\mathcal{L}$ can be computed as follows:
\begin{equation}
\small
  \mathcal{L} = \lambda \mathcal{L}_{\mathrm{MLM}} + (1 - \lambda) \mathcal{L}_{\mathrm{ALM}},
\label{eq:loss}
\end{equation}
where $\lambda$ is a scaling factor between $0$ and $1$.

\noindent
\textbf{Inference.} 
AVLFormer takes video frames and audios as the input for visual and audio encoders and the $[\texttt{BOS}]$ token for the word embedding in inference. It then generates descriptions in an auto-regressive manner. Specifically, it predicts the next word iteratively until the $[\texttt{EOS}]$ token or max sequence length is reached.

%% file: 5_experiments.tex
\begin{figure*}[t]
  \centering
  \vspace{-2mm}
  \includegraphics[width=\textwidth]{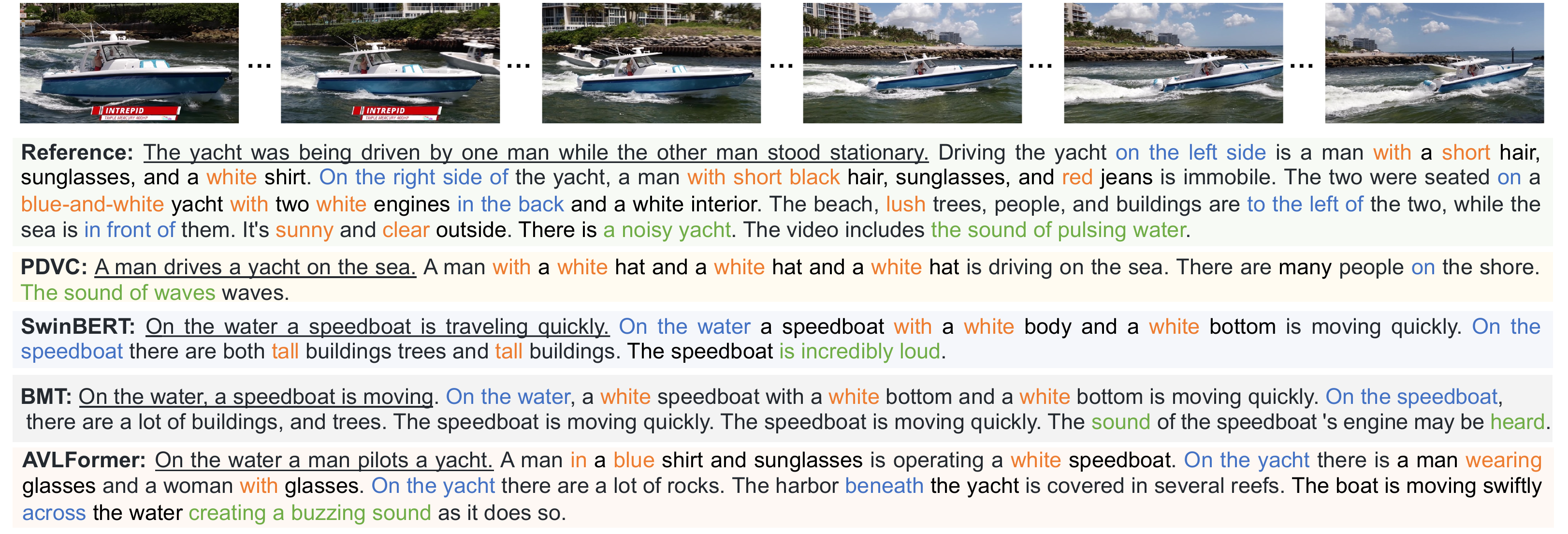}
  \vspace{-7mm}
  \caption{\textbf{Qualitative result of fine-grained video description on FAVDBench.} We compare the proposed AVLFormer with the PDVC~\cite{wang2021end}, SwinBERT~\cite{lin2022swinbert} and BMT~\cite{iashin2020better}. The generated video summary of each method is underlined. The spatial relation, visual adjectives, and audio-related words are highlighted in blue, orange, and green, respectively.}
  \vspace{-5mm}
  \label{fig:favd_performance}
\end{figure*}

\section{Experimental Results}
\label{sec:experiments}

\vspace{-2mm}
\subsection{Implementation details}\label{subsec:implementation} 
\vspace{-2mm}
We conduct experiments on the proposed FAVDBench. All video clips are sampled uniformly to 32 frames, and all frames are resized to the shape of  $224 \times 224$. All audios are sampled by 32kHz in a mono channel. We set the max sequence length and mask probability to 300 and 0.25, respectively for natural language processing. We use the Adam optimizer with a linear warm-up learning rate. The batch size is set to 8 and the maximum number of training epochs is set to 150.

\noindent
\textbf{Evaluation metrics.}\label{subsec:metrics} 
We include 5 conventional captioning metrics as well as the two newly proposed metrics as our evaluation metrics, namely, BLEU~\cite{papineni2002bleu}, Meteor~\cite{denkowski2014meteor}, Rouge-L~\cite{rouge2004package}, CIDEr~\cite{vedantam2015cider}, Clipscore~\cite{hessel2021clipscore}, EntityScore, and AudioScore. We also perform the human evaluation on the generated descriptions. Among the entire test-set, 10$\%$ of samples are randomly selected to be measured, scoring from 0 to 10, and average values of 20 volunteers are reported. The ground truth descriptions are set to 10.
Evaluators are required to focus on the smoothness of descriptions and the correlation between the descriptions and the inputs.

\vspace{-2mm}
\subsection{Main results}
\vspace{-2mm}
\label{subsec:performance_on_favdbench} 
We compare our AVLFormer with the state-of-the-art video captioning method SwinBERT~\cite{lin2022swinbert}, video captioning method with original audio input BMT~\cite{iashin2020better}, dense video captioning method PDVC~\cite{wang2021end} on our FAVDBench for the new FAVD task. Both quantitative and qualitative results are provided in the following sections.

\noindent
\textbf{Quantitative comparison.}\label{subsec:quantitative_comparsion} 
We report the quantitative results over the 8 evaluation metrics in Table~\ref{tab:favd_performance}. We compare AVLFormer with PDVC, SwinBERT and BMT under different backbone freezing settings. AVLFormer beats competitors with a clear margin and achieves the best performance when both the visual and audio backbones are unfrozen. Among the 7 automatic evaluation metrics, the EntityScore is trending closer to human evaluation, demonstrating the validity of the evaluation metric.

\noindent
\textbf{Qualitative comparison.}\label{subsec:qualitative_comparsion} 
We illustrate the qualitative results of AVLFormer and other competitors in Fig.~\ref{fig:favd_performance}. AVLFormer provides finer-grained details of entities and their status than other methods, especially for the spatial location descriptions. For example, AVLFormer attempts to capture the relative positions of major objects, such as the spatial relations among the harbor, the reefs and the yacht, and generates descriptions that are close to the reference. For the audio descriptions, AVLFormer describes the sound and the sounding objects whereas PDVC and SwinBERT often neglect the sounding sources.

\vspace{-2mm}
\subsection{Ablations}
\vspace{-2mm}

\begin{table}[t]\small
\caption{\textbf{Impact of video and audio.} Four different inference modes from top to bottom: positive pairs of video and audio, negative pairs of video and audio, random audio, random video.} 
\vspace{-3mm}
\centering
\label{tab:core_component}
\small
 \begin{tabular}{p{0.8cm}<{\centering}p{0.8cm}<{\centering}p{1.3cm}<{\centering}p{1.5cm}<{\centering}p{1.5cm}<{\centering}}
\toprule[0.8pt]\noalign{\smallskip}
\multicolumn{2}{c}{Inference}        & CIDEr  & EntityScore & AudioScore \\ 
\noalign{\smallskip} \cmidrule(r){1-2} \cmidrule(r){3-5}  
\multicolumn{2}{c}{Pos. V.$\&$A.}    & \textbf{29.85}  & \textbf{44.46}      &  \textbf{63.88}     \\
\multicolumn{2}{c}{Neg. V.$\&$A.}    & 18.63  & 8.25       &  31.25     \\
\multicolumn{2}{c}{Video w Random}   & 20.12  & 40.77      &  61.22     \\
\multicolumn{2}{c}{Random w Audio}   & 1.58   & 11.86      &  44.70     \\
\toprule[0.8pt]
\end{tabular}
\vspace{-4mm}
\end{table}

\begin{table}[t]\small
\caption{\textbf{Audio impact on AudioScore.} Four different audio-visual-text pairs are typed from top to bottom: original, audio-shuffled, audio random initialized and audio fixed initialized pairs.} 
\vspace{-3mm}
\centering
\label{tab:novel_metrics}
\small
 \begin{tabular}{p{1.8cm}<{\centering}p{1.8cm}<{\centering}p{2.5cm}<{\centering}p{2.5cm}<{\centering}}
\toprule[0.8pt]\noalign{\smallskip}
\multicolumn{2}{c}{Pair Type}        & Top-1 AudioScore  & Top-2 AudioScore   \\ 
\noalign{\smallskip} \cmidrule(r){1-2} \cmidrule(r){3-4} 

\multicolumn{2}{l}{Type I}    & \textbf{63.88}  & \textbf{64.60}         \\
\multicolumn{2}{l}{Type II}   &   52.59         &  56.31         \\
\multicolumn{2}{l}{Type III}  &   41.93       &  38.06      \\
\multicolumn{2}{l}{Type IV}   &   39.31         &  40.12         \\
\toprule[0.8pt]
\end{tabular}
\vspace{-7mm}
\end{table}

\begin{figure}[t]
  \centering
  \vspace{0mm}
  \includegraphics[width=0.48\textwidth]{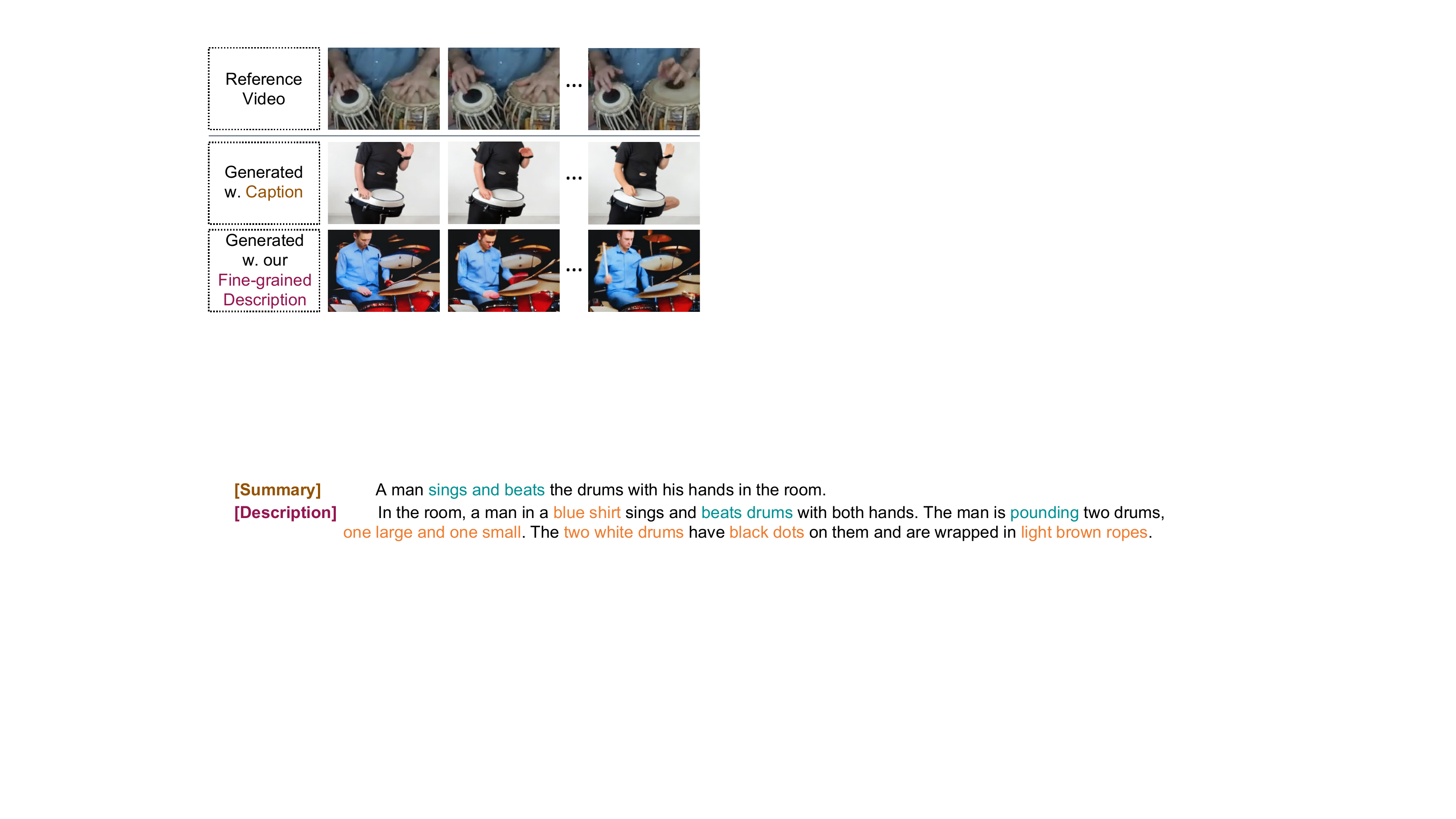}
  \vspace{-7mm}
  \caption{\textbf{Qualitative examples of video generation on FAVDBench.} We use Cogvideo~\cite{hong2022cogvideo} to generate videos through the caption and our fine-grained descriptions
  We eliminate audio-related input descriptions as current video generation models cannot produce sound. \textit{Caption}: ``A man sings and beats the drums with his hands in the room.''; \textit{Description}: ``In the room, a man in a blue shirt sings and beats drums with both hands. The man is pounding two drums, one large and one small. The two white drums have black dots on them and are wrapped in light brown ropes.''} 
  \vspace{-3mm}
  \label{fig:video_generation}
\end{figure}

\begin{figure}[t]
  \centering
  \includegraphics[width=0.48\textwidth]{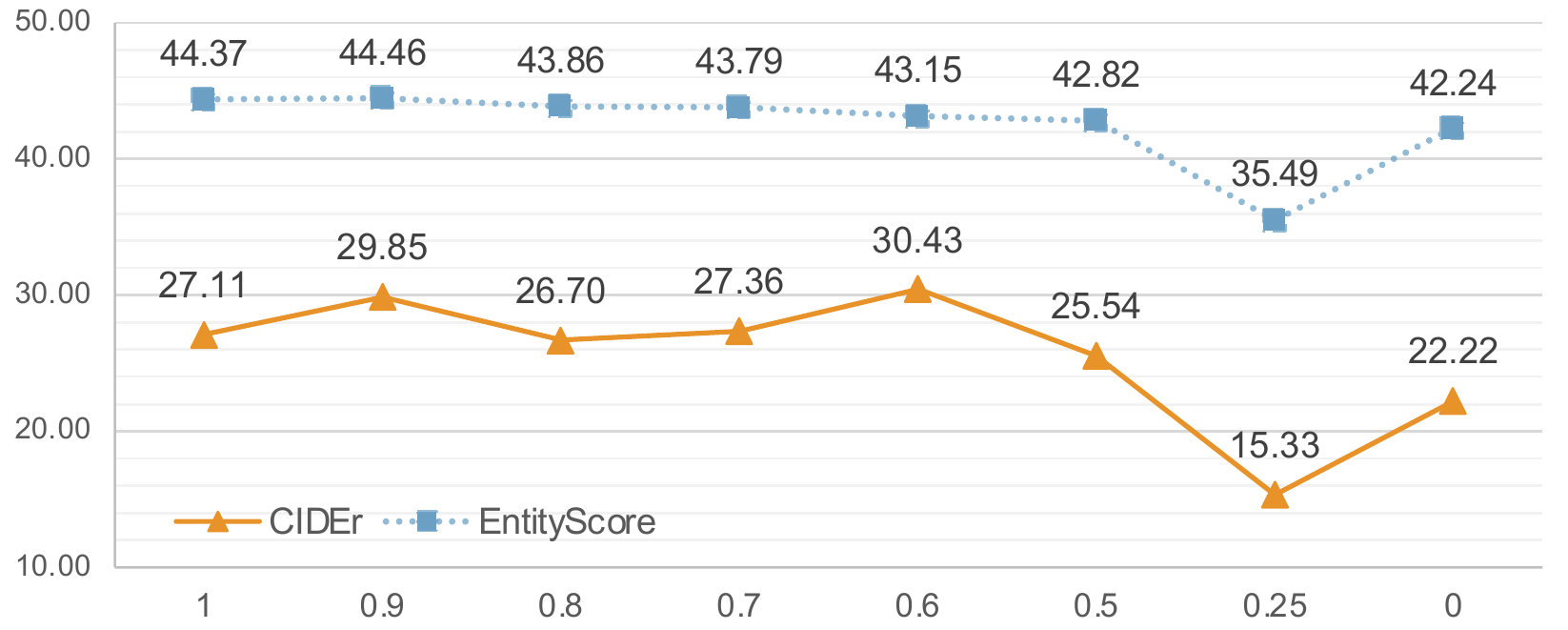}
  \vspace{-7mm}
  \caption{\textbf{Analysis of losses.} We report model performance from on and EntityScore with respect to different $\lambda$ values of loss.}
  \vspace{-7mm}
  \label{fig:lambda_loss}
\end{figure}

We first ablate the impact of audio and visual modalities on AVLFormer by replacing the positive audio-visual pairs with negative pairs or random noises in inference. The results are shown in Table~\ref{tab:core_component}. We then analyze the contribution of the masked language modeling loss and the auto-regressive language modeling loss in Fig.~\ref{fig:lambda_loss}. Finally, we evaluate the validity of our proposed metrics.

\noindent
\textbf{Impact of audio.}\label{subsec:audio_impact}
We empirically find that the trained model can infer partial sound-related descriptions without an audio branch as shown in Fig.~\ref{fig:favd_performance} and Table~\ref{tab:core_component}. It is probably because audio is often related to actions, and the network can predict sound descriptions by understanding the actions. However, in this situation, the sound descriptions are often incorrect or unrelated to the actual audio. For example, the model generates one-third of samples with a pattern like: \emph{somebody/something makes noise}.

\noindent
\textbf{Impact of visual.}\label{subsec:video_impact}
As shown in Table~\ref{tab:core_component}, replacing visual information with random noises will cause a significant performance drop in all metrics. This indicates that in FAVD, the visual modality is most important. Even with the aid of correct audio information, the model cannot produce related descriptions without visual guidance.

\noindent
\textbf{Analysis of losses.}

We employ both the masked language modeling loss $\mathcal{L}_{\text{MLM}}$ and the auto-regressive language modeling loss $\mathcal{L}_{\text{ALM}}$ as our objective loss functions. A ratio $\lambda$ is adopted to balance the weights between them (Eq.~\ref{eq:loss}). We train our model with different $\lambda$s under the same setting and plot the test set accuracy in Fig.~\ref{fig:lambda_loss}. As shown, using pure $\mathcal{L}_{\text{MLM}}$ (\ie, $\lambda = 1$) achieves significant better performance than using pure $\mathcal{L}_{\text{ALM}}$ (\ie, $\lambda = 0$), \eg, 27.11 \emph{vs} 22.22 in CIDEr and 44.37 \emph{vs} 42.24 in EntityScore. The model achieves the best performance when $\lambda = 0.9$.

\noindent
\textbf{Analysis of proposed metrics.}
The EntityScore indicates how comprehensively described entities are when compared to reference descriptions. As shown in Table~\ref{tab:favd_performance} and Table~\ref{tab:core_component}, the EntityScore displays a scoring pattern that is fairly similar to human evaluation, which demonstrates the validity of the EntityScore. For the AudioScore, we additionally design an ablation to evaluate its validity. Specifically, we compare four settings of audio-visual-text pairs: Type I: the original audio-visual-text pairs; Type II: shuffle the audios while keeping the visual-text pairs unchanged; Type III: replace the audio with uniformly random initialized noise; Type IV: replace the audio with a fixed 1e-3 value initialization. The results are shown in Table~\ref{tab:novel_metrics}. Type I keeping the semantic-consistent audio and visual-text pairs achieves significantly higher AudioScore than others, proving the validity of the AudioScore.

\vspace{-1mm}
\subsection{Video generation}
\vspace{-2mm}

Existing video generation models can also benefit from our FAVDBench as we provide finer-grained video descriptions than captions. To prove our claim, we compare the generated videos from captions and our fine-grained description using a state-of-the-art text-video model Cogvideo~\cite{hong2022cogvideo}. As shown in Fig.~\ref{fig:video_generation}, the video generated from the descriptions is closer to the reference video than the one generated from the captions. Furthermore, FAVDBench can be used to train video generation models and can substantially reduce the impact of insufficient scene description on video generation tasks as mentioned in~\cite{singer2022make}.

%% file: 6_conclusion.tex
\vspace{-1mm}
\section{Conclusion}
\vspace{-2mm}
We proposed FAVD, a new audio-visual-language modeling task that aims to generate paragraph-level fine-grained textual descriptions for every object in the given audible videos. A new benchmark called FAVDBench was constructed to facilitate the research. Two new metrics were designed to evaluate the quality of the generated descriptions, where the EntityScore is used to assess the completeness of entities in descriptions and the AudioScore evaluates the accuracy of audio descriptions. We also presented a transformer-based architecture AVLFormer for this task. Extensive experiments validated our model and metrics design. We intend to 
explore architectures in FAVDBench~\cite{qin2022cosformer,sun2022vicinity,qin2023toeplitz}, and use FAVDBench to investigate audible video generation in the future.

\noindent
\textbf{Acknowledgement:}
We thank Lei Li and Hai Jie for their valuable advice on this work. This work is partially supported by the National Key R\&D Program of China (NO.2022ZD0160100) and partly by the Shanghai Committee of Science and Technology (Grant No. 21DZ1100100).

%% file: 7_appendix.tex
\section{Appendix}

In the appendix, we additionally provide the word frequency of FAVDBench and more ablation study results such as the impact of different attention masking strategies in AVLFormer and extended quantitative results on FAVDBench. We also include more video generation samples as well as their quantitative results. A video (\href{https://youtu.be/iWJvTB-bTWk}{https://youtu.be/iWJvTB-bTWk}) containing some samples in FAVDBench is attached to this supplementary material. 

\subsection{The FAVDBench}
\label{sec:appendix_dataset}

\begin{figure}[t]
  \centering
  \vspace{1mm}
  \includegraphics[width=0.48\textwidth]{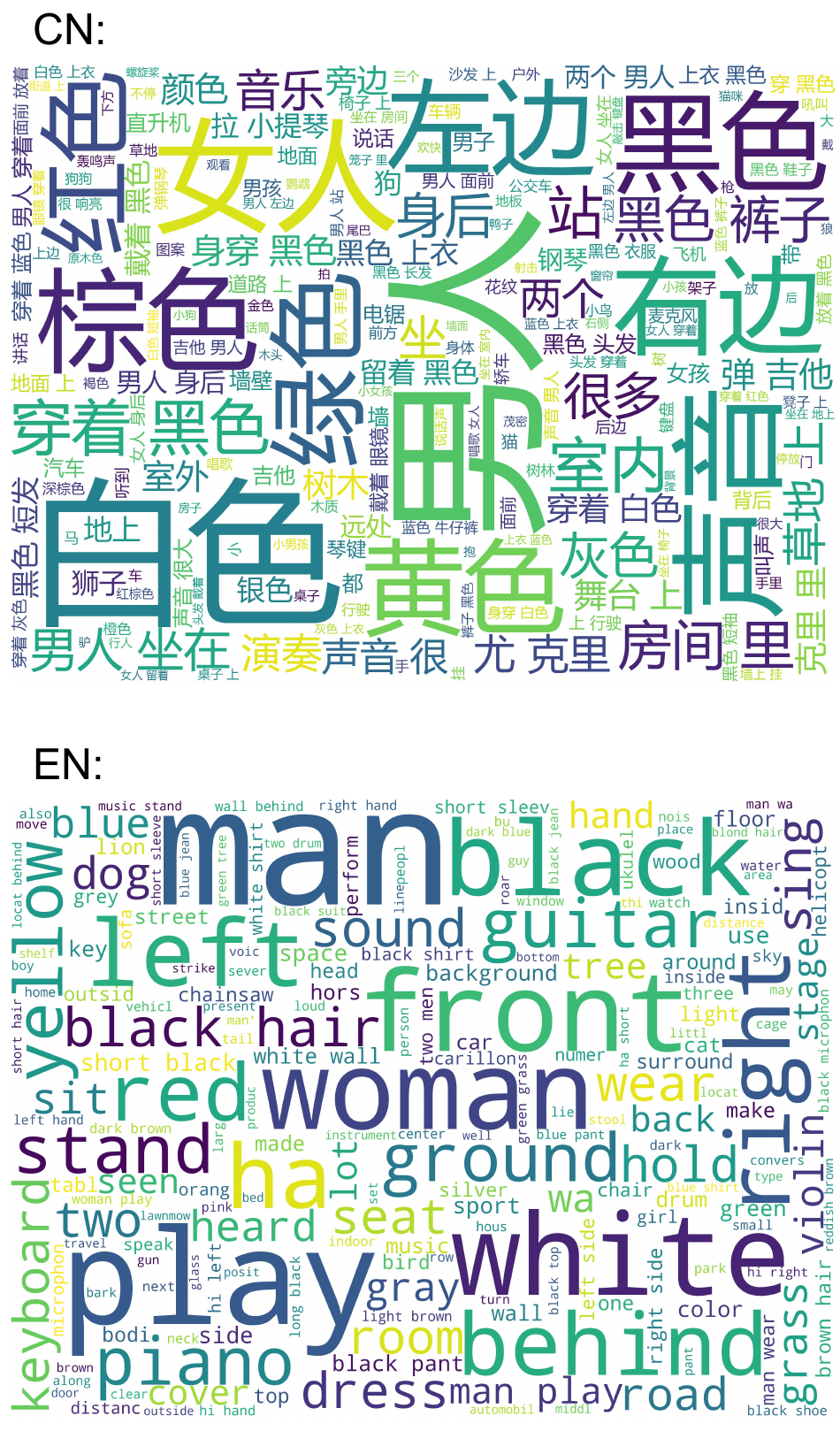}
  \caption{\textbf{Word Frequency.} We count the word frequency of Chinese annotation (upper) and English annotation (bottom) on FAVDBench.}
  \label{fig:word_freq}
\end{figure}

\noindent
\textbf{Word Frequency.}
As FAVDBench provides both English and Chinese descriptions, we plot the frequency of English and Chinese vocabularies in Fig.~\ref{fig:word_freq}. The most common words in descriptions, such "man" and "woman," are nouns. Whereas, both adjectives and prepositions also appear in the most frequent vocabularies of descriptions (\eg, ``red'', ``black'', ``left'', ``right''). It is worth noting that sound-related vocabularies occur in the word frequency diagram, such as ``sound''.  It proves that our descriptions of videos thoroughly capture the visual and audio details, including actions, characteristics, and relative spatial relations. 

\subsection{Experimental Results}
\label{sec:appendix_experiments}

\begin{figure*}[t]
    
    \small
    \centering
    \label{tab:att_mask}
    \small
    \begin{tabular}{p{2cm}<{\centering}p{2cm}<{\centering}p{1.6cm}<{\centering}p{1.6cm}<{\centering}p{1.7cm}<{\centering}p{1.7cm}<{\centering}p{1.6cm}<{\centering}p{1.8cm}<{\centering}p{1.8cm}<{\centering}}
    \toprule[0.8pt]\noalign{\smallskip}
    
    \multicolumn{2}{c}{Attention Types}                               & B@1             & B@4             & METEOR            & ROUGE\_L       & CIDEr          & Clipscore      & EntityScore     \\ 
    \noalign{\smallskip} \cmidrule(r){1-2} \cmidrule(r){3-9}  
    \multicolumn{2}{c}{Type \uppercase\expandafter{\romannumeral1}}   & 44.10           & 10.29           & 18.36             & \textbf{30.93} &\textbf{29.85}  & 69.74          & 42.71           \\
    \multicolumn{2}{c}{Type \uppercase\expandafter{\romannumeral2}}   & 19.72           & 1.97            & 8.94              & 23.03          & 0.89           & 60.25          & 20.06            \\
    \multicolumn{2}{c}{Type \uppercase\expandafter{\romannumeral3}}   & 44.44           & 10.24           & 18.39             & 30.65          & 29.21          & 69.91          & 44.30            \\
    \multicolumn{2}{c}{Type \uppercase\expandafter{\romannumeral4}}   & \textbf{44.45}  & 10.41           & 18.47             & 30.75          & 29.16          & \textbf{70.19} & \textbf{44.58}   \\
    \multicolumn{2}{c}{Type \uppercase\expandafter{\romannumeral5}}   & 44.32           & \textbf{10.50}  & \textbf{18.51}    & 30.86          & 29.07          & 70.06          & 44.41            \\
    \toprule[0.8pt]
    \end{tabular}

  \centering
  \vspace{3mm}
  \includegraphics[width=\textwidth]{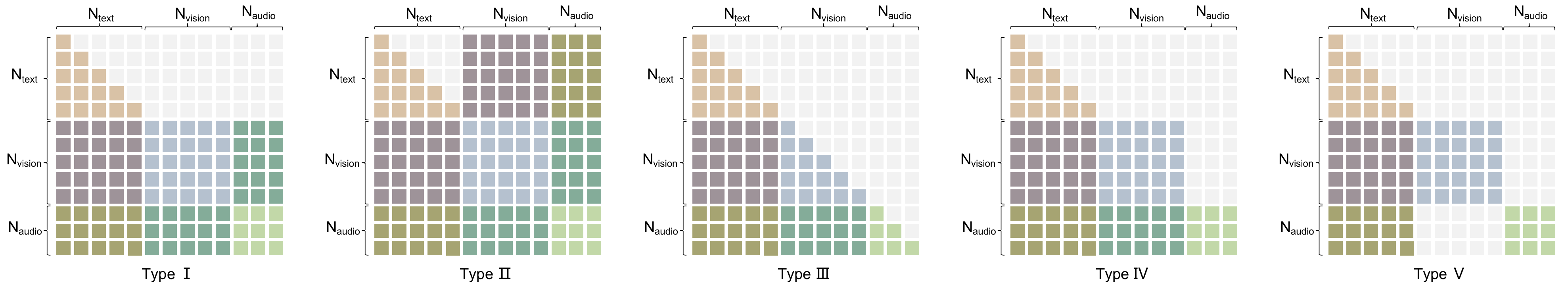}
  \caption{\textbf{Impact of transformer attention masks.} 5 different types of attention masks in the table are visualized in bottom figure. The masked attention is colored gray. The attention masks of text, vision, and audio are colored brown, blue, and green, respectively. The type \uppercase\expandafter{\romannumeral1} is set as the default of AVLFormer, where all vision and audio tokens are visible to text, and vision tokens and audio tokens are visible to each other. The type \uppercase\expandafter{\romannumeral2} is full attention among three modalities. Apart from the type \uppercase\expandafter{\romannumeral1}, all text tokens are visible to vision and audio tokens. The type \uppercase\expandafter{\romannumeral3} is single-direction attention, where the tokens in preceding order are visible to preceding tokens, and vice versa. In type \uppercase\expandafter{\romannumeral4}, vision tokens are visible to audio, and vice versa. In type \uppercase\expandafter{\romannumeral5}, vision tokens and audio tokens are independent.}
  \label{fig:attention_mask}
\end{figure*}

\noindent
\textbf{Impact of different AVLFormer attention masks.}
We ablate the impact of different AVLFormer attention masks in Fig.~\ref{fig:attention_mask}. The empirical results illustrate that the full attention mask (Type \uppercase\expandafter{\romannumeral2}) negatively affects model coverage. It may cause by information leakage when all word tokens are visible to vision and audio tokens when we infer the descriptions in auto-regressive manner. Audio tokens are not visible to vision tokens (Type \uppercase\expandafter{\romannumeral3} - Type \uppercase\expandafter{\romannumeral5}) contributes to performance improvement, compared to the default attention mask (Type \uppercase\expandafter{\romannumeral1}).

\noindent
\textbf{Extended quantitative results.}
We report 6 additional metrics together with the 8 metrics from the main paper in Table~\ref{tab:more_results}. We extend the Clipscore~\cite{hessel2021clipscore} from the first frame to 32 frames. To obtain the later value, all frames are required to measure the cosine similarity with descriptions, and report the the averaged scores. 
Over the 14 evaluation metrics, AVLFormer leads a substantial performance improvement than PDVC and SwinBERT under various backbone freezing settings.

\noindent
\textbf{Quantitative results of SwinBERT.}
We evaluate SwinBERT over 7 representative metrics through loading various pretrained weights and fine-tuning on FAVDBench in Table~\ref{tab:swinbert_performance}. Pretrained weights from other captioning dataset contribute to the performance improvement, especially the VATEX. As a comparison, AVLFormer beats all settings of SwinBERT over all evaluation metrics.

\subsection{Video Generation}
\label{sec:appendix_video_generation}

\noindent
\textbf{Quantitative results.}
To compare the performance between the caption-generated and the description-generated videos, We report the quantitative results over Frechet Video Distance (FVD)~\cite{unterthiner2018towards} metric in Table~\ref{tab:video_generation}. We employ Cogvideo~\cite{hong2022cogvideo} to generate the video based on captions and our fine-grained descriptions. The FVD score reflects that our fine-grained description can generate videos that are more close to the referenced videos.

\begin{table}[t]\small
\caption{\textbf{Video generation performance on FAVDBench.} FVD~\cite{unterthiner2018towards} is used to measure the generated video from captions and our fine-grained descriptions. As the text-video model is used to directly infer without fine-tuning, the score is higher than normal ranges. Besides, the smaller FVD is the better.} 
\centering
\label{tab:video_generation}
\small
    \begin{tabular}{p{1.8cm}<{\centering}p{2.0cm}<{\centering}p{2.8cm}<{\centering}p{3.0cm}<{\centering}}
    \toprule[0.8pt]\noalign{\smallskip}
    
                                      & Captions  & \multicolumn{2}{c}{Fine-grained Descriptions}   \\  \noalign{\smallskip} \midrule
    FVD (\textbf{\textdownarrow})     & 1,540,010  & \multicolumn{2}{c}{\textbf{1,457,802}}         \\
    \toprule[0.8pt]
    \end{tabular}
\end{table}

\begin{figure*}[t]
  \centering
  \includegraphics[width=0.98\textwidth]{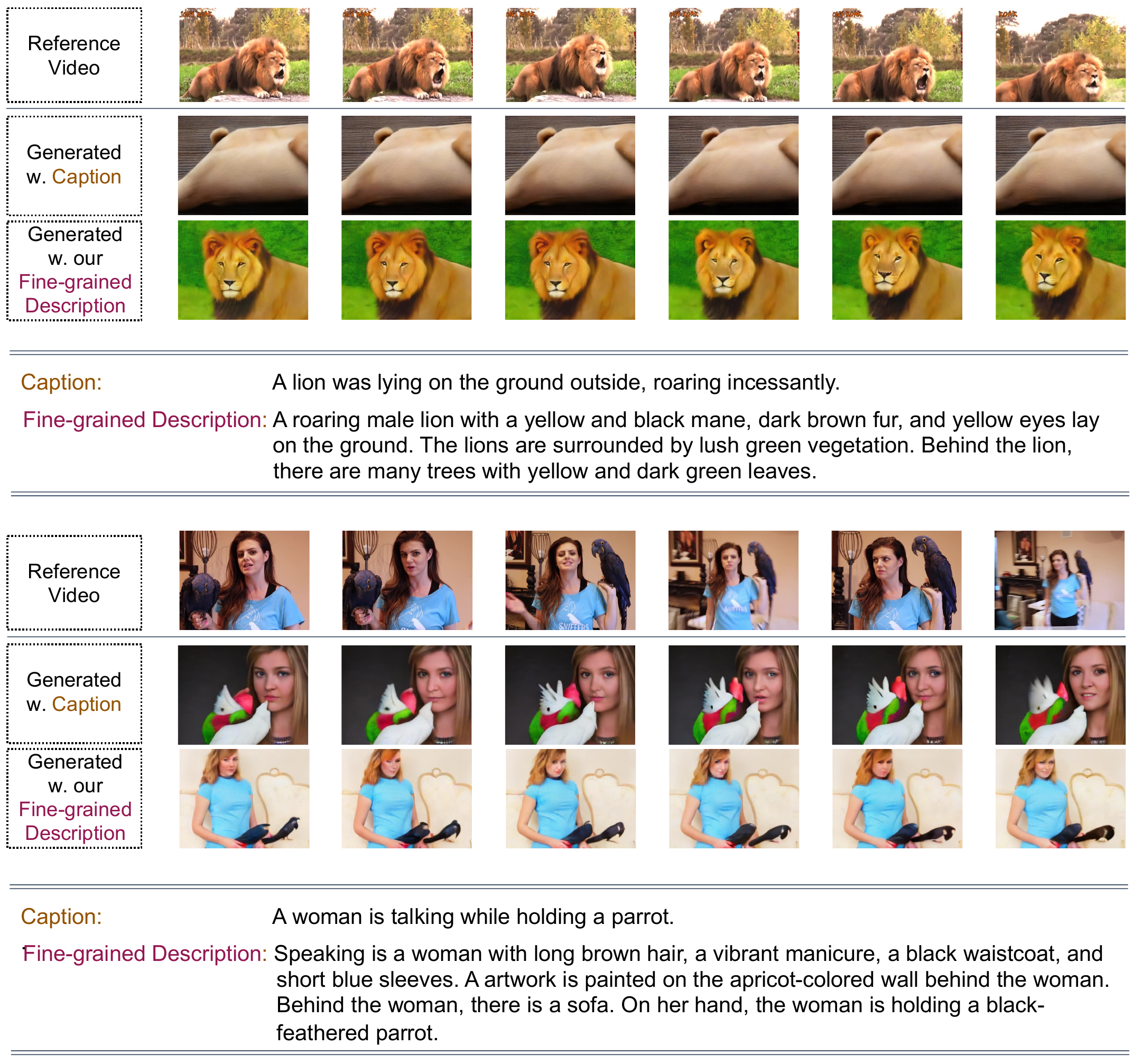}
  \vspace{-2mm}
  \caption{\textbf{More qualitative examples of video generation on FAVDBench (1).} The images in each group are sampled from the ground-truth videos and videos produced by Cogvideo through the caption and our fine-grained descriptions, respectively. } 
  \label{fig:vid_generation_1}
\end{figure*}

\begin{figure*}[t]
  \vspace{-50mm}
  \centering
  \includegraphics[width=0.98\textwidth]{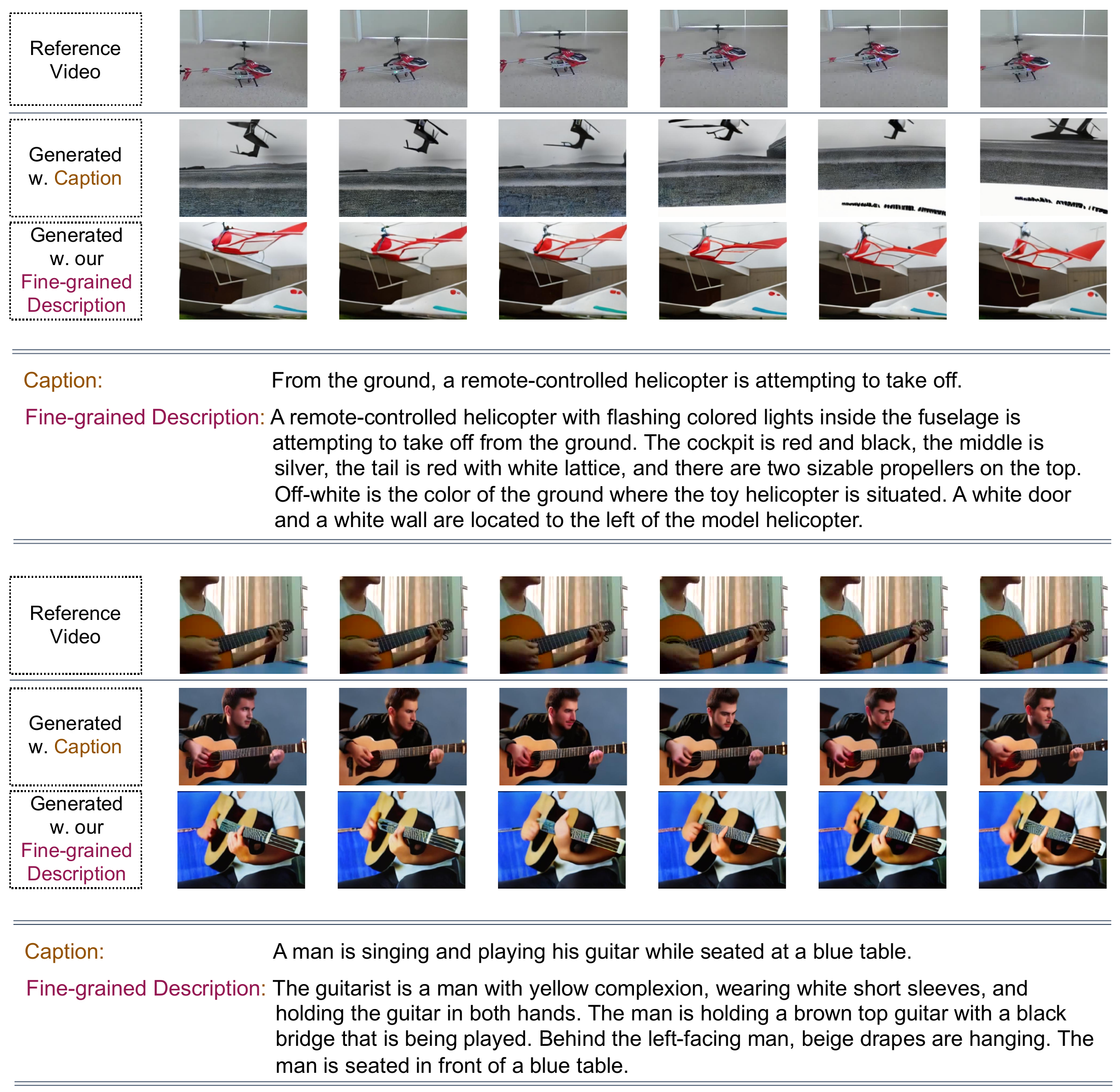}
  \caption{\textbf{More qualitative examples of video generation on FAVDBench (2).} The images in each group are sampled from the ground-truth videos and videos produced by Cogvideo through the caption and our fine-grained descriptions, respectively. } 
  \label{fig:vid_generation_2}
\end{figure*}

\noindent
\textbf{Qualitative results.}
In Fig.~\ref{fig:vid_generation_1} and Fig.~\ref{fig:vid_generation_2}, we provide more video generation examples in our FAVDBench. The videos are generated from captions and our fine-grained descriptions through a state-of-the-art text-video model Cogvideo~\cite{hong2022cogvideo}. FAVDBench can guide video generation more accurately than captions, generating videos that are closer to the reference videos.

\begin{table*}[t]\small
\caption{\textbf{More quantitative comparisons with different methods on FAVDBench.} We report different backbone settings for PDVC, SwinBERT and AVLFormer. In the backbone freeze (FRZ.) column, \textbf{-} represents that input is not available; \ding{220} represents that is trainable; \ding{91} represents frozen. As an extension of Table 2 in the main paper, we report Bleu-1, Bleu-2, Bleu-3, Bleu-4, METEOR, ROUGE\_L, CIDEr, Clipscore 1 frame, Reference Clipscore 1 frame, Clipscore 32 frames (in average), Reference Clipscore 32 frames (in average), EntityScore, and AudioScore. Scores of human evaluation are not repeated in this table. For all metrics, higher values are better. }
\centering
\label{tab:more_results}

\begin{tabular}{p{2.4cm}<{\centering}p{.7cm}<{\centering}p{.7cm}<{\centering}p{1.1cm}<{\centering}p{1.1cm}<{\centering}p{1.1cm}<{\centering}p{1.1cm}<{\centering}p{1.5cm}<{\centering}p{1.5cm}<{\centering}p{1.5cm}<{\centering}}
\toprule[0.8pt]
\multirow{2}{*}{Method}  & \multicolumn{2}{c}{Backbone FRZ.} & \multicolumn{7}{c}{Conventional Metric}   \\ \cmidrule(r){2-3} \cmidrule(r){4-10}
                    & Visual          & Audio         & B@1            & B@2            & B@3             & B@4            & METEOR         & ROUGE\_L       & CIDEr          \\ \midrule
PDVC~\cite{wang2021end}                & \ding{220}      & \textbf{-}    & 34.91          & 19.90          & 11.15           & 6.33           & 13.80          & 24.67          & 6.27           \\ 
PDVC w. Audio       & \ding{220}      & \ding{220}    & 35.59          & 20.47          & 11.53           & 6.47           & 14.49          & 26.98          & 13.13          \\ \midrule
SwinBERT~\cite{lin2022swinbert}            & \ding{220}      & \textbf{-}    & 39.68          & 23.54          & 14.26           & 9.05           & 16.78          & 29.60          & 21.69          \\
SwinBERT~\cite{lin2022swinbert}            & \ding{91}       & \textbf{-}    & 41.41          & 24.43          & 14.57           & 9.08           & 17.19          & 29.49          & 23.03          \\ \midrule
AVLFormer           & \ding{220}      & \ding{220}    & \textbf{44.10} & \textbf{26.61} & \textbf{16.21}  & \textbf{10.29} & \textbf{18.36} & \textbf{30.93} & \textbf{29.85}  \\
AVLFormer           & \ding{91}       & \ding{220}    & \textbf{44.10} & 26.39          & 16.02           & 10.23          & 18.25          & 30.54          & 26.31           \\
AVLFormer           & \ding{220}      & \ding{91}     & 42.82          & 25.79          & 15.80           & 10.16          & 17.96          & 30.68          & 25.45            \\
AVLFormer           & \ding{91}       & \ding{91}     & 42.35          & 25.37          & 15.40           & 9.84           & 17.73          & 30.25          & 25.65            \\
\toprule[0.8pt] 
\end{tabular}

\begin{tabular}{p{2.4cm}<{\centering}p{.7cm}<{\centering}p{.7cm}<{\centering}p{1.6cm}<{\centering}p{1.6cm}<{\centering}p{1.6cm}<{\centering}p{1.6cm}<{\centering}p{1.5cm}<{\centering}p{1.5cm}<{\centering}}
\toprule[0.8pt]
\multirow{2}{*}{Method}  & \multicolumn{2}{c}{Backbone FRZ.} & \multicolumn{4}{c}{Clipscore} & \multicolumn{2}{c}{Proposed Metric}   \\ \cmidrule(r){2-3} \cmidrule(r){4-7} \cmidrule(r){8-9}
                    & Visual          & Audio         & Fr@1           & Ref. Fr@1      & Fr@32           & Ref. Fr@32     & EntityScore    & AudioScore              \\ \midrule
PDVC~\cite{wang2021end}                & \ding{220}      & \textbf{-}    & 65.85          & 66.33          & 65.77           & 66.27          & 32.40          & 29.37                   \\ 
PDVC w. Audio       & \ding{220}      & \ding{220}    & 66.35          & 67.16          & 66.15           & 67.04          & 33.09          & 40.33                    \\ \midrule
SwinBERT~\cite{lin2022swinbert}            & \ding{220}      & \textbf{-}    & 68.64          & 70.02          & 68.13           & 69.75          & 38.17          & 34.57                   \\
SwinBERT~\cite{lin2022swinbert}            & \ding{91}       & \textbf{-}    & 68.39          & 70.02          & 67.98           & 69.79          & 39.39          & 35.06                   \\ \midrule
AVLFormer           & \ding{220}      & \ding{220}    & 70.24          & \textbf{72.09} & 69.74           & \textbf{71.82} & \textbf{42.70} & \textbf{52.43}  \\
AVLFormer           & \ding{91}       & \ding{220}    & 69.42.         & 71.30          & 68.98           & 71.04          & 41.69          & 51.18                   \\
AVLFormer           & \ding{220}      & \ding{91}     & \textbf{70.58} & 71.97          & \textbf{70.07}  & 71.69          & 41.36          & 51.72                  \\
AVLFormer           & \ding{91}       & \ding{91}     & 69.82          & 71.33          & 69.33           & 71.05          & 41.39          & 51.01                   \\
\toprule[0.8pt] 
\end{tabular}

\end{table*}

\begin{table*}[t]\small
\vspace{-3mm}
\caption{\textbf{Comparison with different pretrained weights of SwinBERT~\cite{lin2022swinbert} on FAVDBench.} Pretrained weights from 5 captioning datasets are evaluated: MSR-VTT, MSVD, TVC, VATEX, YOUCOOK \uppercase\expandafter{\romannumeral2}. As a comparison, results of SwinBERT and AVLFormer training with default settings are reported. All pretrained weights are provided by the SwinBERT official repository.}
\vspace{-3mm}
\centering
\label{tab:swinbert_performance}
 \begin{tabular}{p{2.4cm}<{\centering}p{0.7cm}<{\centering}p{0.8cm}<{\centering}p{1.0cm}<{\centering}p{1.0cm}<{\centering}p{1.2cm}<{\centering}p{1.2cm}<{\centering}p{1.5cm}<{\centering}p{1.5cm}<{\centering}p{1.5cm}<{\centering}}
\toprule[0.8pt]
\multirow{2}{*}{Method} & \multicolumn{2}{c}{\multirow{2}{*}{pretrained Weights}} & \multicolumn{5}{c}{Conventional Metric}  & \multicolumn{2}{c}{Proposed Metric}  \\ \cmidrule(r){4-8} \cmidrule(r){9-10}
                      &             &               & B@1            & B@4            & Meteor          & CIDEr          & Clipscore     & EntityScore     & AudioScore        \\ \midrule
SwinBERT              & \multicolumn{2}{c}{\ding{56}}           & 39.68          & 9.05           & 16.78           & 21.69          & 68.13          & 39.60          & 34.57              \\
SwinBERT & \multicolumn{2}{c}{MSR-VTT~\cite{xu2016msr}}         & 42.59          & 9.66           & 17.84           & 23.16          & \textbf{70.40} & 41.41          & 35.11              \\ 
SwinBERT & \multicolumn{2}{c}{MSVD~\cite{chen2011collecting}}   & 42.16          & \textbf{10.15} & 17.89           & 26.85          & 69.69          & 40.91          & 34.48              \\ 
SwinBERT & \multicolumn{2}{c}{TVC~\cite{lei2020tvr}}            & 42.24          & 9.67           & 17.71           & 24.18          & 70.33          & 41.55          & 35.27              \\ 
SwinBERT & \multicolumn{2}{c}{VATEX~\cite{wang2019vatex}}       & \textbf{43.03} & 10.02          & \textbf{18.01}  & 25.59          & 69.74          & \textbf{41.77} & 35.44              \\ 
SwinBERT & \multicolumn{2}{c}{YOUCOOK \uppercase\expandafter{\romannumeral2}~\cite{ZhXuCoAAAI18}} & 42.67          & 9.05           & 17.96           & \textbf{27.64} & 68.13          & 41.76          & \textbf{35.45}     \\ \midrule
AVLFormer (Ref.)     & \multicolumn{2}{c}{\ding{56}}            & \textbf{44.10} & \textbf{10.29} & \textbf{18.36}  & \textbf{29.85} & \textbf{69.74} & \textbf{44.46} & \textbf{52.43}      \\
\toprule[0.8pt]
\end{tabular}
\end{table*}

